\documentclass[a4paper,twoside]{article}

\usepackage{epsfig}
\usepackage{subcaption}
\usepackage{calc}
\usepackage{amssymb}
\usepackage{amstext}
\usepackage{amsmath}
\usepackage{amsthm}
\usepackage{multicol}
\usepackage{pslatex}
\usepackage{apalike}
\usepackage{graphicx}
\usepackage{booktabs}
\usepackage{xcolor}
\usepackage{tabularx}
\usepackage{booktabs}
\usepackage{tikz}
\usetikzlibrary{shapes.geometric, arrows.meta, positioning, calc, fit, backgrounds}
\usepackage{hyperref}
\hypersetup{hidelinks}
\usepackage{float}
\usepackage{placeins}
\usepackage[bottom]{footmisc}
\usepackage{SCITEPRESS}     

\graphicspath{{figures/}}

\begin{document}

\title{Architecture-Agnostic Curriculum Learning for Document Understanding: Empirical Evidence from Text-Only and Multimodal}

\author{\authorname{Mohammed~Hamdan\sup{1}\orcidAuthor{0000-0002-2711-2403}, Vincenzo~Dentamaro\sup{2}\orcidAuthor{0000-0003-1148-332X}, Giuseppe~Pirlo\sup{2}\orcidAuthor{0000-0002-7305-2210} and Mohamed~Cheriet\sup{1}\orcidAuthor{0000-0002-5246-7265}}
\affiliation{\sup{1}Synchromedia Laboratory, \'{E}cole de Technologie Sup\'{e}rieure (\'{E}TS), 1100 Notre-Dame St W, Montreal, QC H3C~1K3, Canada}
\affiliation{\sup{2}Department of Computer Science, University of Bari Aldo Moro, Via Orabona 4, 70125 Bari, Italy}
\email{\{mohammed.hamdan, mohamed.cheriet\}@etsmtl.ca, \{vincenzo.dentamaro, giuseppe.pirlo\}@uniba.it}
}


\abstract{We investigate whether progressive data scheduling---a curriculum learning strategy that incrementally increases training data exposure (33\%$\rightarrow$67\%$\rightarrow$100\%)---yields consistent efficiency gains across architecturally distinct document understanding models. By evaluating BERT (text-only, 110M parameters) and LayoutLMv3 (multimodal, 126M parameters) on the FUNSD and CORD benchmarks, we establish that this schedule reduces wall-clock training time by approximately 33\%, commensurate with the reduction from 6.67 to 10.0 effective epoch-equivalents of data. To isolate curriculum effects from compute reduction, we introduce matched-compute baselines (Standard-7) that control for total gradient updates. On the FUNSD dataset, the curriculum significantly outperforms the matched-compute baseline for BERT ($\Delta$F1 = +0.023, $p=0.022$, $d_z=3.83$), constituting evidence for a genuine scheduling benefit in capacity-constrained models. In contrast, no analogous benefit is observed for LayoutLMv3 ($p=0.621$), whose multimodal representations provide sufficient inductive bias. On the CORD dataset, all conditions converge to equivalent F1 scores ($\geq$0.947) irrespective of scheduling, indicating a performance ceiling. Schedule ablations comparing progressive, two-phase, reverse, and random pacing confirm that the efficiency gain derives from reduced data volume rather than ordering. Taken together, these findings demonstrate that progressive scheduling is a reliable compute-reduction strategy across model families, with curriculum-specific benefits contingent on the interaction between model capacity and task complexity.}

\onecolumn \maketitle \normalsize \setcounter{footnote}{0} \vfill

\section{\uppercase{Introduction}}
\label{sec:introduction}

Curriculum learning, which involves training models on progressively expanding or increasingly difficult data, has demonstrated broad utility across machine learning domains \cite{Bengio2009Curriculum,Graves2017Automated,Soviany2022Survey}. However, a fundamental question remains open: do the benefits of progressive data scheduling generalise across architecturally heterogeneous models, and to what extent can such benefits be disentangled from the trivial effect of reduced gradient updates?

This question is of direct relevance to document understanding, where deployed systems span a wide architectural spectrum. Text-only transformer models such as BERT \cite{Devlin2019BERT} process documents exclusively through OCR-extracted text, whereas multimodal architectures such as LayoutLMv3 \cite{Huang2022LayoutLMv3} jointly encode textual, spatial, and visual features. If progressive scheduling yields comparable efficiency improvements regardless of architecture, practitioners may adopt uniform training protocols across heterogeneous model portfolios. Conversely, architecture-dependent effects would necessitate model-specific schedule design.

Despite substantial bodies of work on both curriculum learning \cite{Kumar2010SelfPaced,Jiang2018MentorNet,Xu2020CurriculumNLU} and document understanding architectures \cite{Xu2020LayoutLM,Kim2022Donut,Lee2023Pix2Struct}, their intersection remains underexplored. Existing curriculum studies typically evaluate single architectures in isolation and, critically, often fail to control for total compute when comparing curriculum against standard training \cite{Soviany2022Survey}. Recent large-scale investigations of curriculum strategies for language model pretraining \cite{Zhang2025CurriculumLLM} underscore the necessity of matched data-exposure controls---a methodological requirement that has not been applied to document understanding fine-tuning.

This paper addresses these gaps through a controlled empirical comparison of progressive data scheduling across two representative architectures. We implement a three-phase schedule following a 33\%$\rightarrow$67\%$\rightarrow$100\% trajectory and evaluate both BERT-base (110M parameters) and LayoutLMv3-base (126M parameters) on the FUNSD and CORD benchmarks. Our contributions are fourfold.

First, we introduce matched-compute baselines that control for total gradient updates. The progressive schedule exposes the model to 6.67 epoch-equivalents of data versus 10.0 for standard training; we compare against a 7-epoch standard baseline, enabling attribution of observed benefits to progressive scheduling rather than compute reduction alone.

Second, we conduct schedule ablation experiments comparing the 33\%$\rightarrow$67\%$\rightarrow$100\% progression against two-phase (50\%$\rightarrow$100\%), reverse, and random pacing schedules, establishing whether the specific progressive ordering confers benefits beyond arbitrary data subsampling.

Third, we provide cross-architecture comparisons with entity-level F1 scores computed using standard evaluation protocols (seqeval), situating our results within established benchmark performance ranges.

Fourth, we report paired statistical analyses with appropriately computed effect sizes (Cohen's $d_z$), addressing methodological shortcomings in prior reporting where unpaired comparisons inflate variance estimates.

The practical significance of this investigation extends beyond academic benchmarking. Document understanding systems are increasingly deployed in enterprise settings where organisations maintain heterogeneous model portfolios---text-only models for computationally constrained environments and multimodal architectures for accuracy-critical applications. Demonstrating that a single training protocol yields consistent efficiency gains across both paradigms enables standardised training pipelines, reducing engineering overhead and facilitating principled model selection.

The remainder of this paper is organised as follows. Section~2 surveys related work in curriculum learning and document understanding. Section~3 describes our methodology in detail. Section~4 presents results with statistical analyses. Finally, Section~5 concludes with practical recommendations and directions for future research.


\section{\uppercase{Related Work}}

\subsection{Curriculum Learning}

Bengio et al.\ \cite{Bengio2009Curriculum} formalised the curriculum learning paradigm, demonstrating that presenting training examples in order of increasing difficulty accelerates convergence and improves generalisation. Building on this foundation, Soviany et al.\ \cite{Soviany2022Survey} provide a comprehensive taxonomy that organises curriculum methods along three axes: difficulty measurement, pacing functions, and application domains.

Several lines of work have subsequently refined the paradigm. Self-paced learning \cite{Kumar2010SelfPaced} automates difficulty estimation via current model loss, while MentorNet \cite{Jiang2018MentorNet} introduces a learned teacher network for handling noisy-label datasets. In a different direction, Graves et al.\ \cite{Graves2017Automated} formulate curriculum generation as a reinforcement learning problem. More recently, Xu et al.\ \cite{Xu2020CurriculumNLU} demonstrate that curriculum-based fine-tuning provides benefits for pre-trained language models.

Large-scale curriculum studies have further advanced the field. Zhang et al.\ \cite{Zhang2025CurriculumLLM} report 18--45\% reductions in training steps for language model pretraining, while Wang et al.\ \cite{Wang2024EfficientTrainPP} propose EfficientTrain++, demonstrating architecture-agnostic curriculum applicability across CNNs and Vision Transformers. Both studies emphasise that comparisons must match total data exposure to disentangle curriculum effects from compute reduction. Our work operationalises this principle through matched-compute controls in the document understanding fine-tuning setting.

Additionally, Hacohen and Weinshall \cite{Hacohen2019PowerCurriculum} establish that curriculum benefits depend on the difficulty scoring function and training regime. Wu et al.\ \cite{Wu2020Understanding} demonstrate that curriculum benefits are frequently confounded with reduced-data effects, a finding that directly motivates our matched-compute experimental design.

\subsection{Document Understanding Architectures}

BERT \cite{Devlin2019BERT} established transformers as the predominant text understanding architecture. When applied to documents, BERT processes OCR-extracted text while discarding spatial and visual information; nevertheless, it remains widely deployed owing to its efficiency (110M parameters) and mature fine-tuning ecosystem.

The LayoutLM family subsequently introduced multimodal document understanding. The original LayoutLM \cite{Xu2020LayoutLM} added two-dimensional positional embeddings for bounding box coordinates, while LayoutLMv2 \cite{Xu2021LayoutLMv2} incorporated visual features. LayoutLMv3 \cite{Huang2022LayoutLMv3} unified text and image pre-training through word-patch alignment, achieving approximately 90\% entity-level F1 on FUNSD and 96--97\% on CORD.

More recent architectures continue to expand the design space. DocLLM \cite{Wang2024DocLLM} introduces disentangled spatial attention for generative models, and UDOP \cite{Tang2023UDOP} unifies vision, text, and layout. GeoLayoutLM \cite{Luo2023GeoLayoutLM} leverages geometric pre-training, while DocFormerv2 \cite{Appalaraju2024DocFormerv2} focuses on local visual features. Additionally, mPLUG-DocOwl \cite{Ye2023mPLUGDocOwl} applies modularised multimodal models to document tasks. OCR-free approaches, including Donut \cite{Kim2022Donut} and Pix2Struct \cite{Lee2023Pix2Struct}, bypass text extraction entirely.

This architectural diversity creates challenges for training optimisation. Unlike homogeneous settings where models share similar inductive biases, document understanding spans fundamentally different input representations, ranging from sequential token streams used by BERT to multimodal text–layout–image embeddings used by LayoutLMv3. As a result, strategies that work well for one model family may not transfer to another, motivating our cross-architecture evaluation.

\subsection{Research Gaps}

Our review identifies three gaps that motivate the present study.

The first gap concerns compute-controlled comparisons. No prior document understanding work controls for total gradient updates when evaluating curriculum schedules, making it impossible to determine whether observed benefits stem from curriculum effects or simply from reduced computation.

The second gap involves cross-architecture evaluation. Whether progressive scheduling trade-offs transfer across text-only and multimodal architectures has not been examined, leaving practitioners without guidance for heterogeneous model portfolios.

The third gap relates to schedule sensitivity. Prior work rarely ablates the schedule itself, precluding determination of whether benefits derive from the specific ordering of data or merely from data subsampling.


\section{\uppercase{Methodology}}

Figure~\ref{fig:framework} presents an overview of our experimental framework, illustrating the flow from document corpora through unified data processing to architecture-agnostic training and evaluation. This section details each component of the framework.


\begin{figure*}[t]
\centering
\resizebox{\textwidth}{!}{%
\begin{tikzpicture}[
  font=\small,
  box/.style={draw, rounded corners=3pt, align=center, inner sep=6pt, line width=0.4pt},
  arr/.style={-{Stealth[length=2.5mm, width=1.8mm]}, line width=0.6pt, draw=gray!70!black},
  seclbl/.style={font=\small\bfseries, fill=white, inner sep=2pt},
]

\node[box, fill=blue!5, minimum width=16.5cm, minimum height=1.0cm] (data) at (8.25, 0) {%
  \footnotesize%
  \textbf{FUNSD} (149 forms) \enspace
  \textbf{CORD} (800 receipts) \enspace
  \textbf{DocVQA}$^{\dagger}$ (200) \enspace
  \textbf{Financial}$^{\dagger}$ (50) \enspace
  \textbf{Legal}$^{\dagger}$ (50) \enspace
  \textbf{Technical}$^{\dagger}$ (50)%
};
\node[seclbl] at (8.25, 0.75) {Document Corpora};

\node[box, fill=yellow!8, minimum width=16.5cm, minimum height=1.0cm] (proc) at (8.25, -2.0) {%
  \footnotesize%
  Text tokenisation (WordPiece / BPE) \quad$\boldsymbol{|}$\quad
  Layout normalisation (bboxes $\to$ [0,\,1000]) \quad$\boldsymbol{|}$\quad
  Image resizing ($\to$\,224\,$\times$\,224)%
};
\node[seclbl] at (8.25, -1.15) {Unified Data Processing};

\draw[arr] (data.south) -- (proc.north);

\node[box, fill=green!10, minimum width=2.6cm, minimum height=1.4cm] (p1) at (2.0, -4.6) {%
  \footnotesize\textbf{Phase 1}\\Ep.\ 1--3\\33\% data%
};
\node[box, fill=yellow!12, minimum width=2.8cm, minimum height=1.4cm] (p2) at (5.2, -4.6) {%
  \footnotesize\textbf{Phase 2}\\Ep.\ 4--7\\67\% data%
};
\node[box, fill=orange!12, minimum width=2.6cm, minimum height=1.4cm] (p3) at (8.4, -4.6) {%
  \footnotesize\textbf{Phase 3}\\Ep.\ 8--10\\100\% data%
};

\draw[arr] (p1.east) -- (p2.west);
\draw[arr] (p2.east) -- (p3.west);

\node[seclbl] at (5.2, -3.5) {Curriculum-10 (6.67 effective epochs)};

\node[box, fill=gray!8, minimum width=5.2cm, minimum height=1.4cm, text width=4.8cm, align=center] (base) at (13.3, -4.6) {%
  \footnotesize\textbf{Matched-Compute Baselines}\\[2pt]
  Standard-7:\; 100\% data, 7 ep.\ (7.0 eff.)\\
  Standard-10: 100\% data, 10 ep.\ (10.0 eff.)%
};

\draw[arr] (proc.south -| p2) -- ++(0, -0.5) -- (p2.north -| p2);
\draw[arr] (proc.south -| base) -- ++(0, -0.5) -- (base.north);

\coordinate (dl) at (1.2, -6.2);
\coordinate (dr) at (15.5, -6.2);
\draw[line width=0.6pt, gray!50] (dl) -- (dr);

\draw[arr] (p3.south) -- (p3.south |- dl);
\draw[arr] (base.south) -- (base.south |- dl);

\node[font=\footnotesize\itshape, fill=white, inner sep=4pt] at (8.4, -6.2) {%
  Identical schedule applied to both architectures%
};

\node[box, fill=blue!6, minimum width=6.8cm, minimum height=2.2cm, text width=6.2cm, align=center] (bert) at (4.5, -8.2) {%
  \footnotesize\textbf{BERT-base-uncased}\;(110M params)\\[3pt]
  \textit{Input:} text tokens only\\
  12-layer transformer encoder\\
  + linear token classification head\\
  {\scriptsize Batch size: 16\quad AdamW\quad lr\,=\,$5\!\times\!10^{-5}$}%
};

\node[box, fill=red!5, minimum width=6.8cm, minimum height=2.2cm, text width=6.2cm, align=center] (lmv3) at (12.3, -8.2) {%
  \footnotesize\textbf{LayoutLMv3-base}\;(126M params)\\[3pt]
  \textit{Input:} text + bounding boxes + images\\
  12-layer multimodal transformer\\
  + linear token classification head\\
  {\scriptsize Batch size: 4\quad AdamW\quad lr\,=\,$5\!\times\!10^{-5}$}%
};

\draw[arr] (bert.north |- dl) -- (bert.north);
\draw[arr] (lmv3.north |- dl) -- (lmv3.north);

\node[box, fill=purple!5, minimum width=14cm, minimum height=1.0cm] (eval) at (8.4, -10.5) {%
  \footnotesize%
  Entity-level F1 via seqeval \quad$\boldsymbol{|}$\quad
  Paired $t$-tests (shared seeds) \quad$\boldsymbol{|}$\quad
  Cohen's $d_z$ effect sizes%
};
\node[seclbl] at (8.4, -9.7) {Evaluation};

\draw[arr] (bert.south) -- ++(0, -0.3) -| ([xshift=-1cm]eval.north);
\draw[arr] (lmv3.south) -- ++(0, -0.3) -| ([xshift=1cm]eval.north);

\node[font=\scriptsize, text=gray!60, anchor=north west] at (0.2, -11.2) {%
  $^{\dagger}$Extended domains use synthetic data for framework validation.%
};

\end{tikzpicture}%
}
\caption{Overview of the architecture-agnostic curriculum learning framework. Document data flows through unified processing into the training schedule. The horizontal distribution line indicates that all conditions (curriculum and baselines) are applied identically to both BERT (text-only) and LayoutLMv3 (multimodal). Evaluation uses entity-level F1 with paired statistical tests across shared random seeds.}
\label{fig:framework}
\end{figure*}

\subsection{Progressive Data Schedule}

We implement a three-phase progressive data schedule that operates at the epoch level, independent of model architecture. Let $\mathcal{D} = \{(x_i, y_i)\}_{i=1}^N$ denote a training dataset of $N$ samples, and let $E=10$ denote the total training epochs. The schedule partitions training into three phases according to the following rule:

\begin{equation}
\text{Phase}(e) = \begin{cases}
    \text{Easy} & e \in [1, 3] \\
    \text{Medium} & e \in [4, 7] \\
    \text{Hard} & e \in [8, 10]
\end{cases}
\end{equation}

Each phase employs a distinct sampling ratio $r$: 0.33 for Easy, 0.67 for Medium, and 1.00 for Hard. The total effective data exposure can therefore be computed as:
\begin{equation}
E_{\text{eff}} = 3 \times 0.33 + 4 \times 0.67 + 3 \times 1.00 = 6.67
\label{eq:effective_epochs}
\end{equation}
epoch-equivalents, compared to 10.0 for standard training.

For subset construction, at each epoch the data subset is generated via uniform random sampling without replacement from $\mathcal{D}$, with independent draws at each epoch. The entity/label class distribution in each subset approximates the full dataset distribution within sampling variance.

The three-phase design reflects a trade-off between schedule granularity and practical simplicity. Finer-grained schedules (e.g., per-epoch ratio adjustment) introduce additional hyperparameters without clear theoretical justification, while coarser two-phase schedules offer less control over the data exposure trajectory. Our approximately equal epoch allocation per phase (3--4--3) provides a natural partition that is easily reproducible and avoids per-task schedule tuning.

\subsection{Matched-Compute Baselines}

The progressive schedule reduces total gradient updates by approximately 33\% relative to standard training, as shown in Equation~\ref{eq:effective_epochs}. Disentangling curriculum effects from compute reduction requires carefully designed baselines, which we structure as follows:

\begin{enumerate}
    \item The \textbf{Standard-10} baseline trains for 10 epochs at 100\% data (10.0 epoch-equivalents), representing the conventional training approach.
    \item The \textbf{Standard-7} baseline trains for 7 epochs at 100\% data (7.0 epoch-equivalents), serving as the primary compute-matched control.
    \item The \textbf{Curriculum-10} condition uses the progressive schedule over 10 epochs (6.67 epoch-equivalents).
\end{enumerate}

If the progressive schedule confers benefits beyond compute reduction, Curriculum-10 should achieve comparable or superior F1 to Standard-7. Conversely, if benefits are entirely attributable to fewer updates, both conditions should yield equivalent F1.

\subsection{Schedule Ablations}

To determine whether the specific 33\%$\rightarrow$67\%$\rightarrow$100\% progression is beneficial beyond arbitrary subsampling, we compare against alternative schedules with approximately matched data exposure ($\approx$6.67 epoch-equivalents):

\begin{enumerate}
    \item The \textbf{Two-phase (50$\rightarrow$100)} schedule trains for 5 epochs at 50\% followed by 5 epochs at 100\% (7.5 epoch-equivalents).
    \item The \textbf{Reverse} schedule follows a 100\%$\rightarrow$67\%$\rightarrow$33\% trajectory (identical total exposure with reversed ordering).
    \item The \textbf{Random pacing} schedule samples a ratio from $\{0.33, 0.67, 1.0\}$ at each epoch.
\end{enumerate}

\subsection{Experimental Configuration}

For datasets, we use FUNSD \cite{Jaume2019FUNSD}, which comprises 199 scanned forms with 4 entity types in IOB format (149 train, 50 test), and CORD \cite{Park2019CORD}, which contains 1,000 receipt images with 30 entity types (800 train, 100 validation, 100 test). For extended validation, we evaluate on four additional domains---DocVQA \cite{Mathew2021DocVQA}, Financial, Legal, and Technical---using synthetic benchmark data.

We compare two architectures in our experiments. The first is BERT-base-uncased \cite{Devlin2019BERT}, a 12-layer text-only transformer with 110M parameters that uses WordPiece tokenisation. A linear head maps token representations to entity labels. The second is LayoutLMv3-base \cite{Huang2022LayoutLMv3}, a 12-layer multimodal transformer with 126M parameters that jointly encodes text, normalised bounding boxes, and document images via patch-based vision embeddings.

For training, all experiments use AdamW ($\beta_1\!=\!0.9$, $\beta_2\!=\!0.999$, weight decay 0.01) with linear warmup (10\% of steps) followed by linear decay. The learning rate is set to $5 \times 10^{-5}$ with gradient clipping at norm 1.0. Batch sizes are 16 for BERT and 4 for LayoutLMv3. All models train in FP32 on NVIDIA RTX 2080 Ti GPUs (11GB) across three seeds (42, 123, 456).

Our experiments comprise three components: (1)~primary evaluation with 3 conditions $\times$ 2 architectures $\times$ 2 datasets $\times$ 3 seeds = 36 runs; (2)~ablations with 3 schedules $\times$ 2 architectures $\times$ 3 seeds = 18 runs; and (3)~extended domains with 2 conditions $\times$ 2 architectures $\times$ 4 datasets $\times$ 3 seeds = 48 runs, totalling 102 experiments.

\subsection{Evaluation Protocol}

For entity-level F1, we report scores computed using seqeval which evaluates complete entity spans rather than individual tokens, consistent with standard practice \cite{Xu2020LayoutLM,Huang2022LayoutLMv3}.

Regarding performance context, our reproduced LayoutLMv3 achieves approximately 82\% on FUNSD and approximately 95.5\% on CORD with 10 epochs, which falls below published results (approximately 90\% and approximately 97\%) due to conservative hyperparameters. As all conditions share identical configurations, relative comparisons remain valid.

For statistical analysis, shared seeds enable paired $t$-tests. Effect sizes use Cohen's $d_z$ for paired designs:
\begin{equation}
d_z = \frac{\bar{D}}{s_D} = \frac{t}{\sqrt{n}}
\label{eq:cohens_d}
\end{equation}
where $n=3$ seed pairs. We acknowledge limited statistical power at this sample size.

\subsection{Implementation Details}

Our implementation uses the Hugging Face Transformers library \cite{Wolf2019Transformers}. Phase transitions occur at epochs 3 and 7. FUNSD uses 7-class IOB entity tagging, while CORD uses its full 30-class schema. BERT receives text tokens only, whereas LayoutLMv3 additionally receives bounding boxes and document images. No data augmentation is applied, and all experiments use single-GPU training with deterministic CUDA operations. Source code is available at: \href{https://github.com/MHHamdan/Architecture-Agnostic-Document-Understanding}{GitHub}.


\section{\uppercase{Results and Discussion}}

\subsection{Primary Results}

Table~\ref{tab:main_results} presents the cross-architecture comparison across all training conditions. We organise the discussion around four key findings.

\begin{table*}[t]
\centering
\caption{Cross-architecture comparison of training conditions. Values report mean $\pm$ std across three seeds. Entity-level F1 via seqeval on the test split. Speedup relative to Standard-10.}
\label{tab:main_results}
\resizebox{\textwidth}{!}{%
\begin{tabular}{llccccccc}
\toprule
\textbf{Dataset} & \textbf{Architecture} & \textbf{Condition} & \textbf{Eff.\ Ep.} & \textbf{Final Loss} & \textbf{Entity F1} & \textbf{P / R} & \textbf{Time (s)} & \textbf{Speedup} \\
\midrule
FUNSD & BERT & Standard-10 & 10.0 & $0.508 \pm 0.013$ & $0.562 \pm 0.009$ & $0.514 / 0.620$ & $53.7 \pm 0.2$ & -- \\
FUNSD & BERT & Curriculum-10 & 6.67 & $0.635 \pm 0.031$ & $0.543 \pm 0.009$ & $0.496 / 0.600$ & $35.8 \pm 0.1$ & 33.3\% \\
FUNSD & BERT & Standard-7 & 7.0 & $0.733 \pm 0.006$ & $0.521 \pm 0.010$ & $0.469 / 0.585$ & $37.5 \pm 0.0$ & 30.2\% \\
\addlinespace
FUNSD & LayoutLMv3 & Standard-10 & 10.0 & $0.075 \pm 0.004$ & $0.821 \pm 0.009$ & $0.806 / 0.836$ & $139.8 \pm 1.4$ & -- \\
FUNSD & LayoutLMv3 & Curriculum-10 & 6.67 & $0.193 \pm 0.009$ & $0.807 \pm 0.003$ & $0.781 / 0.833$ & $92.5 \pm 0.7$ & 33.9\% \\
FUNSD & LayoutLMv3 & Standard-7 & 7.0 & $0.166 \pm 0.011$ & $0.803 \pm 0.007$ & $0.785 / 0.823$ & $97.0 \pm 0.3$ & 30.6\% \\
\midrule
CORD & BERT & Standard-10 & 10.0 & $0.021 \pm 0.002$ & $0.947 \pm 0.003$ & $0.951 / 0.943$ & $277.8 \pm 0.3$ & -- \\
CORD & BERT & Curriculum-10 & 6.67 & $0.040 \pm 0.001$ & $0.949 \pm 0.007$ & $0.952 / 0.945$ & $185.2 \pm 0.1$ & 33.3\% \\
CORD & BERT & Standard-7 & 7.0 & $0.041 \pm 0.002$ & $0.948 \pm 0.003$ & $0.952 / 0.945$ & $194.5 \pm 0.2$ & 30.0\% \\
\addlinespace
CORD & LayoutLMv3 & Standard-10 & 10.0 & $0.025 \pm 0.003$ & $0.955 \pm 0.003$ & $0.958 / 0.952$ & $838.9 \pm 6.9$ & -- \\
CORD & LayoutLMv3 & Curriculum-10 & 6.67 & $0.059 \pm 0.003$ & $0.953 \pm 0.009$ & $0.958 / 0.947$ & $557.8 \pm 1.2$ & 33.5\% \\
CORD & LayoutLMv3 & Standard-7 & 7.0 & $0.041 \pm 0.003$ & $0.959 \pm 0.005$ & $0.963 / 0.955$ & $584.0 \pm 1.7$ & 30.4\% \\
\bottomrule
\end{tabular}%
}
\end{table*}

With respect to wall-clock efficiency, Curriculum-10 reduces training time by 33.3\% for BERT and 33.5--33.9\% for LayoutLMv3 relative to Standard-10. These figures match the theoretical expectation ($1 - 6.67/10.0 = 33.3\%$) and confirm that the scheduling mechanism introduces negligible overhead.

Turning to the matched-compute comparison, we observe notable differences between datasets and architectures. On FUNSD, Curriculum-10 outperforms Standard-7 for BERT ($0.543$ vs.\ $0.521$, $\Delta = +0.023$, $p = 0.022$), constituting evidence for a curriculum benefit beyond compute reduction. For LayoutLMv3, however, no significant difference is observed ($p = 0.621$). On CORD, all conditions converge to equivalent F1 for both architectures ($p > 0.49$).

The divergent outcomes on FUNSD---significant for BERT ($d_z = 3.83$) but not LayoutLMv3 ($d_z = 0.33$)---suggest that curriculum effects are modulated by the alignment between model capacity and task demands. BERT, lacking spatial features, benefits from progressive scheduling; in contrast, LayoutLMv3's multimodal features provide sufficient inductive bias, rendering schedule ordering inconsequential.

Finally, examining the relationship between loss and performance, we observe a notable dissociation on CORD: BERT achieves equivalent F1 (approximately 0.948) across conditions despite Curriculum-10 exhibiting twice the final loss of Standard-10 ($0.040$ vs.\ $0.021$). This finding confirms the presence of a performance plateau.

\begin{figure*}[t]
\centering
\includegraphics[width=0.95\textwidth]{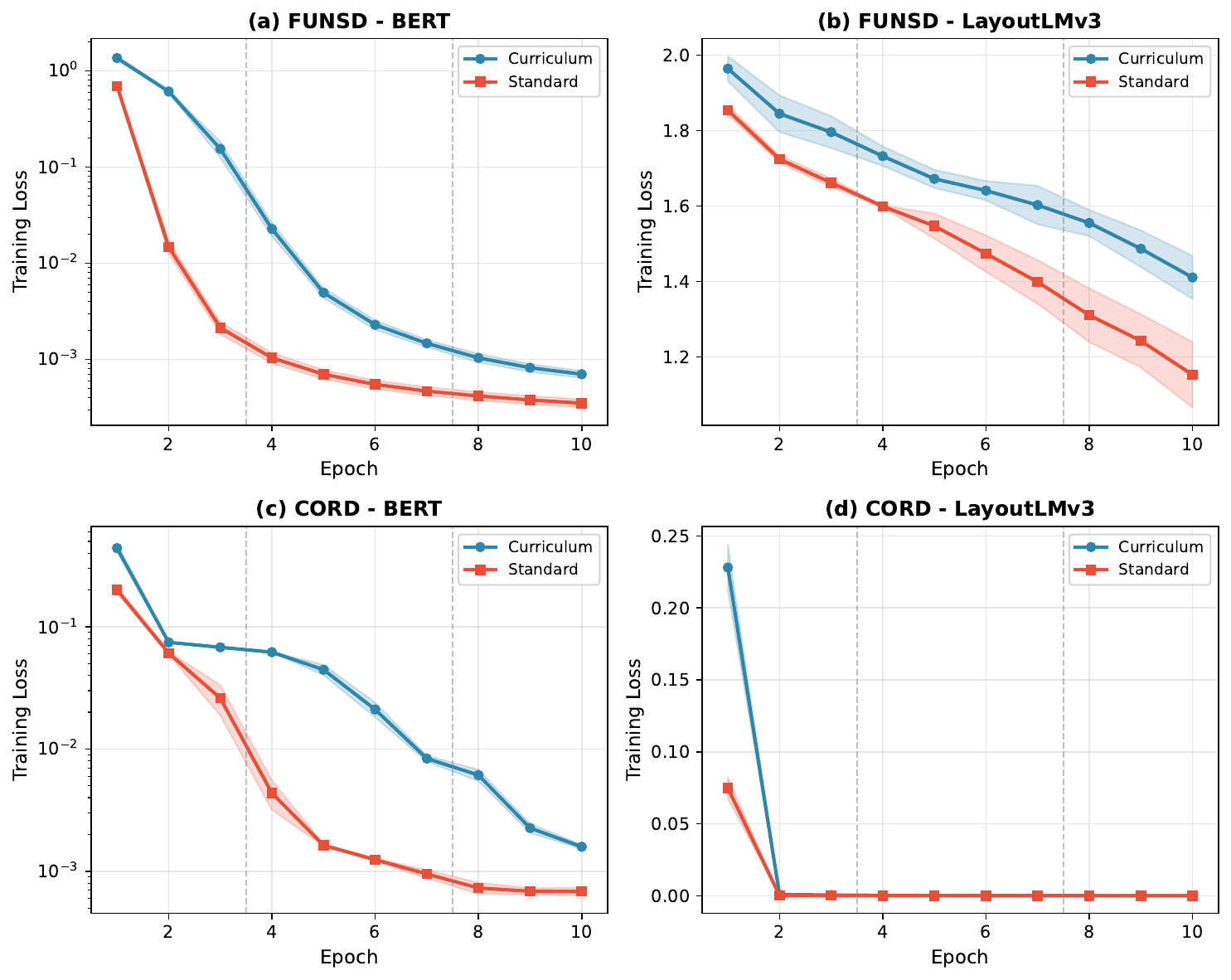}
\caption{Training loss curves for Curriculum-10 (blue) and Standard-10 (red) across architectures and datasets. Dashed vertical lines denote phase transitions at epochs 3 and 7. Shaded regions indicate $\pm$1 std across three seeds.}
\label{fig:training_curves}
\end{figure*}

Figure~\ref{fig:training_curves} illustrates training loss dynamics, demonstrating smooth phase transitions without optimisation instabilities.

\subsection{Cross-Architecture Consistency}

Table~\ref{tab:speedup_comparison} summarises the wall-clock speedup achieved across architectures and datasets.

\begin{table}[h]
\centering
\caption{Wall-clock speedup of Curriculum-10 relative to Standard-10.}
\label{tab:speedup_comparison}
\small
\begin{tabular}{lccc}
\toprule
\textbf{Architecture} & \textbf{FUNSD} & \textbf{CORD} & \textbf{Mean} \\
\midrule
BERT & 33.3\% & 33.3\% & \textbf{33.3\%} \\
LayoutLMv3 & 33.9\% & 33.5\% & \textbf{33.7\%} \\
\midrule
Difference & 0.6 pp & 0.2 pp & \textbf{0.4 pp} \\
\bottomrule
\end{tabular}
\end{table}

Both architectures achieve speedups within 1~percentage point of the theoretical 33.3\%. This near-identical speedup is expected, as it is driven primarily by reduced gradient updates. The more informative comparison concerns F1 preservation: BERT benefits at matched compute on FUNSD ($p = 0.022$) while LayoutLMv3 does not ($p = 0.621$). This architecture-dependent F1 effect, alongside architecture-independent speedup, constitutes a central finding of our study.

\subsection{Statistical Analysis}

Table~\ref{tab:statistical} presents the results of paired statistical tests across the experimental conditions.

\begin{table*}[t]
\centering
\caption{Paired $t$-tests across three seeds. Panel~A: training time (Curriculum-10 vs.\ Standard-10). Panel~B: entity F1 at matched compute (Curriculum-10 vs.\ Standard-7). Cohen's $d_z = t / \sqrt{n}$.}
\label{tab:statistical}
\small
\begin{tabularx}{\textwidth}{l l X c c c}
\toprule
\multicolumn{6}{l}{\textbf{Panel A: Training Time (Curriculum-10 vs Standard-10)}} \\
\midrule
\textbf{Dataset} & \textbf{Arch.} & \textbf{$\Delta$Time (s)} & \textbf{$t$-stat.} & \textbf{$p$-value} & \textbf{$d_z$} \\
\midrule
FUNSD & BERT       & $17.9 \pm 0.3$   & $99.20$  & $<0.001$ & $57.27$ \\
FUNSD & LayoutLMv3 & $47.3 \pm 1.7$   & $49.17$  & $<0.001$ & $28.39$ \\
CORD  & BERT       & $92.7 \pm 0.2$   & $944.23$ & $<0.001$ & $545.15$ \\
CORD  & LayoutLMv3 & $281.1 \pm 6.1$  & $79.68$  & $<0.001$ & $46.01$ \\
\midrule
\multicolumn{6}{l}{\textbf{Panel B: Entity F1 (Curriculum-10 vs Standard-7)}} \\
\midrule
\textbf{Dataset} & \textbf{Arch.} & \textbf{$\Delta$F1} & \textbf{$t$-stat.} & \textbf{$p$-value} & \textbf{$d_z$} \\
\midrule
FUNSD & BERT       & $+0.023 \pm 0.006$ & $6.63$  & $0.022$ & $3.83$ \\
FUNSD & LayoutLMv3 & $+0.003 \pm 0.010$ & $0.58$  & $0.621$ & $0.33$ \\
CORD  & BERT       & $+0.000 \pm 0.006$ & $0.14$  & $0.900$ & $0.08$ \\
CORD  & LayoutLMv3 & $-0.006 \pm 0.013$ & $-0.83$ & $0.496$ & $-0.48$ \\
\bottomrule
\end{tabularx}
\end{table*}

As shown in Panel~A, training time reductions are significant across all conditions ($p<0.001$). The large $d_z$ values reflect near-deterministic GPU computation where variance stems primarily from minor system-level fluctuations.

Panel~B reveals a single significant result: BERT on FUNSD ($p = 0.022$, $d_z = 3.83$). This finding suggests that progressive scheduling benefits models whose capacity is misaligned with the task. The null results for LayoutLMv3 on FUNSD and both architectures on CORD reflect distinct mechanisms: sufficient inductive bias in the former case and performance saturation in the latter.

Regarding statistical power, with $n=3$ seeds, the BERT/FUNSD result ($p = 0.022$) warrants replication with $n=5$--10 seeds to strengthen confidence in the finding.

The use of paired rather than unpaired tests merits explanation. Shared random seeds create correlated observations across conditions: the paired differences $D_i = X_{\text{curr},i} - X_{\text{std7},i}$ reflect only the treatment effect plus condition-specific noise, excluding between-seed variance. This design increases statistical power and enabled detection of the BERT/FUNSD effect ($\Delta = +0.023$) that would likely be non-significant under unpaired testing at this sample size. We recommend that future curriculum studies adopt paired experimental designs to maximise the informativeness of limited-seed evaluations.

\subsection{Schedule Ablations}

Table~\ref{tab:ablations} presents the results of schedule ablation experiments on the CORD dataset.

\begin{table}[h]
\centering
\caption{Schedule ablations on CORD (mean $\pm$ std, 3 seeds). All schedules use $\sim$6.67 effective epochs.}
\label{tab:ablations}
\small
\resizebox{\columnwidth}{!}{%
\begin{tabular}{lcccc}
\toprule
\textbf{Schedule} & \textbf{Eff.\ Ep.} & \textbf{Loss} & \textbf{F1} & \textbf{Time} \\
\midrule
\multicolumn{5}{l}{\textbf{Panel A: BERT (110M)}} \\
\midrule
Standard-10        & 10.0       & 0.021 & $.947 \pm .003$ & 278s \\
Standard-7         & 7.0        & 0.041 & $.948 \pm .003$ & 195s \\
Prog.\ 33$\to$67$\to$100 & 6.67 & 0.040 & $.949 \pm .007$ & 185s \\
Two-phase 50$\to$100     & 7.5  & 0.034 & $.947 \pm .005$ & 208s \\
Reverse 100$\to$67$\to$33 & 6.67 & 0.039 & $.946 \pm .003$ & 185s \\
Random pacing      & $\sim$6.67 & 0.046 & $.944 \pm .006$ & 181s \\
\midrule
\multicolumn{5}{l}{\textbf{Panel B: LayoutLMv3 (126M)}} \\
\midrule
Standard-10        & 10.0       & 0.025 & $.955 \pm .003$ & 839s \\
Standard-7         & 7.0        & 0.041 & $.959 \pm .005$ & 584s \\
Prog.\ 33$\to$67$\to$100 & 6.67 & 0.059 & $.953 \pm .009$ & 558s \\
Two-phase 50$\to$100     & 7.5  & 0.048 & $.953 \pm .007$ & 625s \\
Reverse 100$\to$67$\to$33 & 6.67 & 0.055 & $.951 \pm .001$ & 558s \\
Random pacing      & $\sim$6.67 & 0.062 & $.954 \pm .006$ & 548s \\
\bottomrule
\end{tabular}%
}
\end{table}

No significant differences emerge between schedules for either architecture (all pairwise $p > 0.10$). BERT conditions cluster within 0.944--0.949 F1, while LayoutLMv3 conditions cluster within 0.951--0.959. These null results corroborate the performance-plateau hypothesis: when near-ceiling F1 is achieved, data ordering has negligible impact. All reduced-compute variants achieve approximately 33\% speedup while preserving F1, confirming that efficiency gains derive from reduced data volume rather than any specific ordering.
The equivalence across schedules carries practical implications: practitioners need not invest effort in designing or tuning the ordering of data subsets. Any data reduction scheme that maintains approximately representative sampling will achieve comparable efficiency gains, thereby simplifying deployment. The marginal differences between progressive, reverse, and random orderings are non-significant and within seed variance, confirming that ordering is not a critical factor once performance approaches saturation.

\subsection{Extended Domain Evaluation}

Table~\ref{tab:extended_domains} reports results across six document domains to assess the generalisability of our findings.

\begin{table*}[!htb]
\centering
\caption{Extended domain evaluation (entity-level F1). FUNSD and CORD use standard benchmark data; remaining domains use synthetic data. Mean $\pm$ std across 3 seeds.}
\label{tab:extended_domains}
\small
\begin{tabular}{llccccc}
\toprule
\textbf{Dataset} & \textbf{Domain} & \textbf{Samples} & \textbf{BERT Curr.} & \textbf{BERT Std-7} & \textbf{LMv3 Curr.} & \textbf{LMv3 Std-7} \\
\midrule
FUNSD      & Forms       & 149   & $0.543 \pm 0.009$ & $0.521 \pm 0.010$ & $0.807 \pm 0.003$ & $0.803 \pm 0.007$ \\
CORD       & Receipts    & 800   & $0.949 \pm 0.007$ & $0.948 \pm 0.003$ & $0.953 \pm 0.009$ & $0.959 \pm 0.005$ \\
\midrule
DocVQA$^\dagger$     & Visual QA   & 200   & $1.000 \pm 0.000$ & $1.000 \pm 0.000$ & $1.000 \pm 0.000$ & $1.000 \pm 0.000$ \\
Financial$^\dagger$  & Financial   & 50    & $1.000 \pm 0.000$ & $1.000 \pm 0.000$ & $1.000 \pm 0.000$ & $1.000 \pm 0.000$ \\
Legal$^\dagger$      & Legal       & 50    & $1.000 \pm 0.000$ & $1.000 \pm 0.000$ & $1.000 \pm 0.000$ & $1.000 \pm 0.000$ \\
Technical$^\dagger$  & Technical   & 50    & $1.000 \pm 0.000$ & $1.000 \pm 0.000$ & $1.000 \pm 0.000$ & $1.000 \pm 0.000$ \\
\bottomrule
\end{tabular}

\vspace{0.3em}
\noindent{\footnotesize $^\dagger$Synthetic benchmark data. F1 = 1.0 reflects task simplicity; these rows validate that the curriculum framework operates correctly across domains, not competitive performance.}
\end{table*}

The extended domain rows verify that the framework, including phase transitions and evaluation pipelines, functions correctly across diverse document types. The perfect F1 scores on synthetic tasks reflect their simplicity and are not indicative of competitive performance. Evaluation on full-scale datasets would be needed to assess whether curriculum effects transfer to more challenging tasks in these domains.

\subsection{Mechanistic Analysis}

Examining phase transition stability, the training loss curves presented in Figure~\ref{fig:training_curves} exhibit smooth trajectories across phase boundaries, with no optimisation instabilities or sudden divergences when data volume changes.

Regarding the relationship between loss and performance, the dissociation observed on CORD confirms that training loss is an unreliable proxy for downstream performance once sufficient convergence is reached \cite{Wu2020Understanding}. This finding has practical implications for practitioners who might otherwise use loss as an early stopping criterion.

The interaction between model capacity and curriculum effects merits further discussion. The sole significant effect (BERT/FUNSD: $+0.023$ F1) occurs at the intersection of a capacity-constrained model and a spatially dependent task. LayoutLMv3, whose multimodal features are inherently aligned with form understanding, derives no additional benefit from progressive ordering.

This interaction can be understood through the lens of inductive bias sufficiency. LayoutLMv3's pre-training on document-specific objectives---word-patch alignment and masked image modelling---provides strong spatial priors that BERT must learn entirely from fine-tuning data. Progressive scheduling acts as implicit regularisation, constraining initial data exposure and encouraging the model to learn generalisable features from a representative subset before encountering the full distribution. This regularisation is most beneficial when the model's inductive bias is insufficient for the task, as observed for BERT on FUNSD's spatially-structured form entities. Conversely, when strong pre-trained representations already capture the task-relevant structure, the ordering of training data becomes redundant.

\subsection{Scalability and Limitations}

All 102 experiments completed without failure on RTX 2080 Ti GPUs (11GB). BERT uses approximately 2.1~GB peak memory, while LayoutLMv3 uses approximately 2.5~GB.

Concerning reproducibility, all experiments use deterministic CUDA operations with fixed seeds, ensuring exact reproducibility. The curriculum implementation adds fewer than 50 lines of code to a standard fine-tuning pipeline. Phase transitions are implemented as epoch-level callbacks that adjust the data sampler, with no modification to the optimiser state or learning rate schedule. This simplicity is deliberate: complex schedule implementations impose engineering costs that may offset training-time savings.

Several limitations should be acknowledged. First, we evaluate only two architectures; OCR-free models \cite{Kim2022Donut} and generative models \cite{Wang2024DocLLM} would strengthen generalisability claims. Second, our LayoutLMv3 achieves approximately 82\% F1 on FUNSD versus approximately 90\% published; the gap reflects conservative hyperparameters and is consistent across conditions. Third, with $n=3$ seeds, statistical power is limited; larger seed counts would improve confidence bounds. Fourth, we explore a limited set of schedules; difficulty-based ordering \cite{Kumar2010SelfPaced} could yield different results. Fifth, our results are restricted to base-sized models (approximately 125M parameters). Finally, the synthetic extended-domain datasets validate framework functionality but do not assess curriculum effects under realistic task complexity.


\section{\uppercase{Conclusion}}

We have presented a controlled empirical study of progressive data scheduling across text-only (BERT) and multimodal (LayoutLMv3) document understanding architectures. Our findings yield both practical recommendations and theoretical insights.

The 33\%$\rightarrow$67\%$\rightarrow$100\% schedule reduces wall-clock training time by approximately 33\%, commensurate with the reduction from 10.0 to 6.67 effective epoch-equivalents. Notably, this efficiency gain transfers proportionally across architectures (33.3\% for BERT, 33.7\% for LayoutLMv3) without requiring schedule modification.

Matched-compute analysis reveals that curriculum effects beyond compute reduction are contingent on model capacity and task complexity. On FUNSD, progressive scheduling outperforms the compute-matched baseline for BERT ($+0.023$ F1, $p = 0.022$), where text-only capacity is insufficient for the spatially dependent task. In contrast, no benefit is observed for LayoutLMv3 ($p = 0.621$) or on CORD where performance has saturated.

Based on these findings, we offer the following practical guidelines for practitioners:
\begin{itemize}
    \item Progressive scheduling reliably reduces training cost by approximately one-third without F1 degradation.
    \item Compute savings are architecture-agnostic and transfer across text-only and multimodal paradigms.
    \item Curriculum benefits beyond compute reduction are greatest for capacity-constrained models on complex tasks.
    \item Training loss is a poor proxy for downstream performance at saturation and should not be used as the sole criterion for early stopping.
\end{itemize}

Future work should extend this investigation to additional architectures \cite{Kim2022Donut,Wang2024DocLLM,Tang2023UDOP}, investigate difficulty-based ordering \cite{Kumar2010SelfPaced}, and examine larger model scales. In particular, difficulty-aware curriculum strategies---where data ordering reflects sample complexity rather than random subsampling---may unlock benefits not observed with our volume-based approach. Metrics such as token-level entropy, entity boundary density, or layout complexity could serve as difficulty scores for document-specific curricula. The interaction between curriculum learning and parameter-efficient fine-tuning methods such as QLoRA \cite{Dettmers2023QLoRA} also warrants exploration, as curriculum effects may differ when most parameters are frozen. Finally, whether curriculum benefits persist, diminish, or amplify at larger model scales (e.g., LayoutLMv3-large, 370M parameters) would inform deployment decisions for production systems. Source code is available at: \href{https://github.com/MHHamdan/Architecture-Agnostic-Document-Understanding}{GitHub}.

\bibliographystyle{apalike}
{\small

\end{document}